\documentclass{article}
\usepackage{cite}
\usepackage{newtxtext}
\usepackage{PRIMEarxiv}
\usepackage[utf8]{inputenc} 
\usepackage[T1]{fontenc}    
\usepackage{hyperref}       
\usepackage{url}            
\usepackage{booktabs}       
\usepackage{amsfonts}       
\usepackage{nicefrac}       
\usepackage{microtype}      
\usepackage{lipsum}
\usepackage{fancyhdr}       
\usepackage{graphicx}       
\graphicspath{{media/}}     
\usepackage{amsmath}
\usepackage{algorithm}
\usepackage{algpseudocode}
\usepackage{tikz}
\usepackage{caption}
\usepackage{subcaption}
\usetikzlibrary{positioning, shapes.geometric}
\title{\textbf{Recurrent Expansion: A Pathway Toward the Next Generation of Deep Learning} \\[1ex]
\large \textit{A Brief Technical Note}}

\author{
Tarek Berghout,\href{https://orcid.org/0000-0003-4877-4200}{\includegraphics[height=10pt]{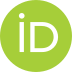}} \\
University of Batna 2 \\
Batna 05000, Algeria \\
\texttt{t.berghout@univ-batna2.dz}
}

\begin{document}
\maketitle

\begin{abstract}

This paper introduces Recurrent Expansion (RE) as a new learning paradigm that advances beyond conventional Machine Learning (ML) and Deep Learning (DL). While DL focuses on learning from static data representations, RE proposes an additional dimension: learning from the evolving behavior of models themselves. RE emphasizes multiple mappings of data through identical deep architectures and analyzes their internal representations (i.e., feature maps) in conjunction with observed performance signals such as loss. By incorporating these behavioral traces, RE enables iterative self-improvement, allowing each model version to gain insight from its predecessors. The framework is extended through Multiverse RE (MVRE), which aggregates signals from parallel model instances, and further through Heterogeneous MVRE (HMVRE), where models of varying architectures contribute diverse perspectives. A scalable and adaptive variant, Sc-HMVRE, introduces selective mechanisms and scale diversity for real-world deployment. Altogether, RE presents a shift in DL: from purely representational learning to behavior-aware, self-evolving systems. It lays the groundwork for a new class of intelligent models capable of reasoning over their own learning dynamics, offering a path toward scalable, introspective, and adaptive artificial intelligence. A simple code example to support beginners in running their own experiments is provided in Code Availability Section of this paper.

\end{abstract}

\keywords{Artificial Intelligence \and Machine Learning \and Deep Learning \and Recurrent Expansion}

\section{Introduction}

ML has fundamentally transformed the way systems learn from data by modeling relationships between inputs and outputs~\cite{mienye2024deepreview, suh2025statisticalDL, kim2025tsf, gheewala2025recsys, talaei2023deepoverview}.
Traditionally, ML tasks are framed as the problem of learning a function as shown in \autoref{eq:ml_basic}. This formulation defines the core of supervised learning, where the goal is to approximate a mapping from inputs to outputs based on observed data, where \( x \) represents input data and \( y \) is the target output.

\begin{equation}
y = f(x)
\label{eq:ml_basic}
\end{equation}

DL, as a powerful subfield of ML, extends this idea by learning complex, layered representations of the input data. This can be formulated as shown in \autoref{eq:dl_formulation}. In \autoref{eq:dl_formulation} In \( x \in \mathcal{X} \) is the raw input data, \( \phi(x) \) represents a sequence of nonlinear transformations applied to \( x \) (i.e., Typically implemented as feature maps extracted by deep neural networks) and \( f \) is the final prediction function that maps the learned representation to the output \( y \in \mathcal{Y} \).

\begin{equation}
y = f(\phi(x))
\label{eq:dl_formulation}
\end{equation}

\begin{figure}[H]
\centering
\resizebox{\linewidth}{!}{
\begin{tikzpicture}[
  box/.style={
    draw, rounded corners=10pt, 
    minimum width=5cm, minimum height=2.2cm, 
    align=center, font=\small,
    text width=4.5cm
  },
  ]

\node[box, thick] (dl) {Deep Learning};

\node[box, above left=1cm and 3cm of dl] (methods) 
  {1. Model Methods: CNN, RNN, Transformer, etc.};
\node[box, above right=1cm and 3cm of dl] (paradigms) 
  {2. Learning Paradigms: Supervised, RL, etc.};
\node[box, below left=1cm and 3cm of dl] (challenges) 
  {3. Technical Challenges: Uncertainty, Interpretability, etc.};
\node[box, below right=1cm and 3cm of dl] (applications) 
  {4. Applications: Classification, Regression, Clustering, Segmentation, etc.};

\draw[<-, thick] (methods) |- (dl);
\draw[<-, thick] (paradigms) |- (dl);
\draw[<-, thick] (challenges) |- (dl);
\draw[<-, thick] (applications) |- (dl);

\end{tikzpicture}
}
\caption{The four foundational pillars of deep learning: model methods, learning paradigms, technical challenges, and practical applications.}
\label{fig:dl_pillars}
\end{figure}
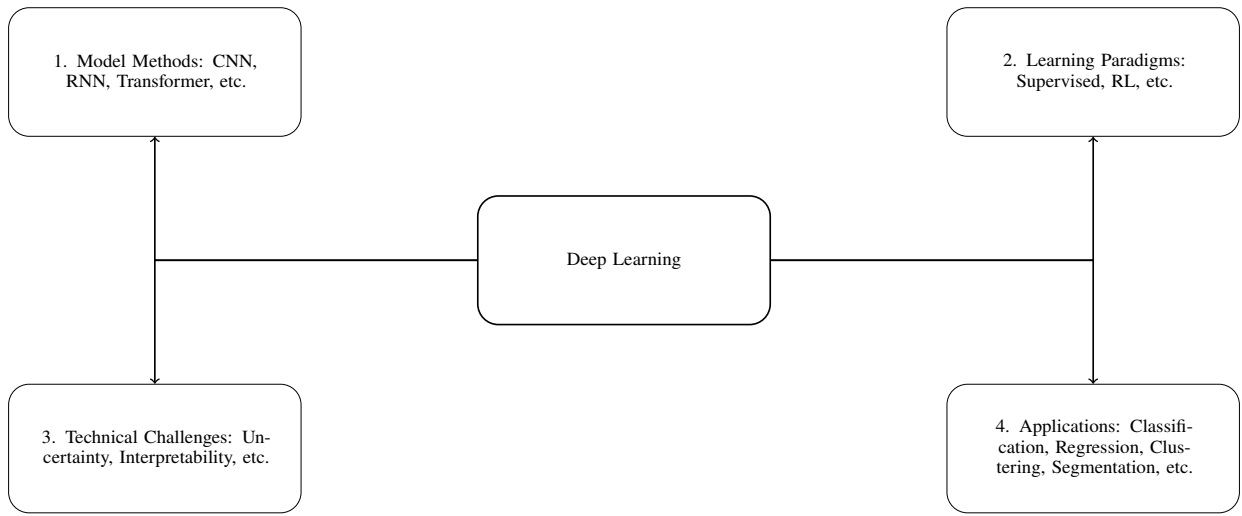

These representations allow models to extract abstract, hierarchical patterns, leading to remarkable success across domains such as vision, language, and control. The field of DL can be broadly categorized along four major pillars: model methods, learning paradigms, technical challenges, and application domains. \autoref{fig:dl_pillars} illustrates the four foundational pillars that structure the DL landscape: model methods, learning paradigms, technical challenges, and applications.

In terms of methods, DL has introduced a variety of architectural innovations designed to model different types of data and tasks. Key model families include Convolutional Neural Networks (CNNs) \cite{Zhao2024} for spatial data such as images, Recurrent Neural Networks (RNNs) and Long Short-Term Memory (LSTM) networks for sequential data \cite{ghojogh2023recurrentneuralnetworkslong}, and Generative Adversarial Networks (GANs) for data synthesis and unsupervised generation\cite{Barsha2025}. More recently, Graph Neural Networks (GNNs) have emerged for processing graph-structured data, and Transformer-based models have revolutionized sequence modeling in both natural language processing and computer vision \cite{Khemani2024}. Each of these methods defines a unique mechanism for learning data representations, either through convolution, recurrence, attention, or adversarial objectives. DL methods are applied under various learning paradigms, each targeting different data availability scenarios and learning objectives \cite{mienye2024deepreview, suh2025statisticalDL, kim2025tsf, talaei2023deepoverview}. The most established paradigm is supervised learning, where models are trained on labeled datasets. Unsupervised learning, in contrast, aims to learn patterns from unlabeled data. Semi-supervised learning combines both labeled and unlabeled data to improve performance when labeled examples are scarce. Reinforcement learning focuses on sequential decision making via rewards and penalties. In addition to these, advanced paradigms such as transfer learning, few-shot learning, federated learning, and self-supervised learning have been introduced to improve generalization, reduce reliance on large labeled datasets, and preserve data privacy. These paradigms define how learning occurs, what supervision is available, and how knowledge can be transferred across tasks or domains.

Despite these innovations, DL still faces critical challenges that limit its broader applicability and trustworthiness \cite{ABDAR2021243, ANTAMIS2024128204}. Among them are uncertainty in predictions, variability across domains or datasets, and heterogeneity in data distributions. Furthermore, DL models often lack interpretability and transparency, making it difficult to understand or explain their decisions. The black-box nature hinders their use in high-stakes applications like healthcare, finance, or autonomous systems. DL underpins a wide range of applications across industries and scientific fields \cite{mienye2024deepreview, suh2025statisticalDL, kim2025tsf, talaei2023deepoverview}. Core tasks include classification (e.g., image recognition), regression (e.g., time-series forecasting), clustering (e.g., customer segmentation), and segmentation (e.g., medical imaging or autonomous driving). These tasks serve as foundational objectives across various domains such as healthcare, natural language processing, robotics, bioinformatics, and recommender systems. Each application imposes unique demands on the model’s architecture, interpretability, and data efficiency.

A fundamental limitation of current approaches is their heavy dependence on learning from data representations alone, without considering another valuable source of information: the behavior of the models themselves. We argue that understanding how models behave across different training runs, tasks, or data transformations can unlock new dimensions of learning. This motivates the concept of RE, a novel framework that builds on model behavior as a self-reflective mechanism to guide future learning and knowledge acquisition.

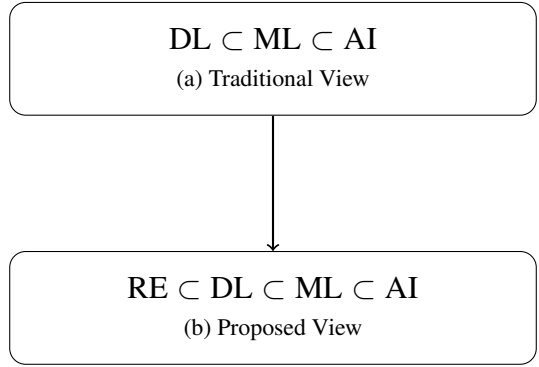
\begin{figure}[ht]
\centering
\begin{tikzpicture}[
    box/.style={draw, rounded corners=6pt, minimum width=7cm, minimum height=1.5cm, font=\large, align=center}
]

\node[box] (trad) {%
    $\text{DL} \subset \text{ML} \subset \text{AI}$\\
    \small (a) Traditional View
};

\node[box, below=1.8cm of trad] (prop) {%
    $\text{RE} \subset \text{DL} \subset \text{ML} \subset \text{AI}$\\
    \small (b) Proposed View
};

\draw[->, thick] (trad.south) -- (prop.north);

\end{tikzpicture}
\caption{Comparison between traditional and proposed perspectives in the artificial intelligence hierarchy.}
\label{fig:hierarchy_view}
\end{figure}

\autoref{fig:hierarchy_view} contrasts the traditional view of artificial intelligence hierarchies with the proposed perspective, where RE is explicitly situated within DL.

The remainder of this paper is organized as follows. Section~\ref{sec:re_framework} introduces the core concept of RE and its foundational mechanism. Section~\ref{sec:mvre} extends this idea to MVRE, incorporating multiple prior models for behavioral feedback. Sections~\ref{sec:hmvre} and~\ref{sec:sc_hmvre} generalize MVRE to heterogeneous and scalable settings, enhancing diversity and adaptability. After that Section \ref{sec:example} follows with and ilustrative example while its codes are made availble in Section \ref{par:code_availability}. In Section~\ref{sec:related}, we position our work within the context of related literature and discuss important future directions. Finally, Section~\ref{sec:conclusion} summarizes the key contributions of the paper.

\section{Recurrent Expansion}
\label{sec:re_framework}

We define RE as a learning framework that builds upon the behavior of previous DL models, introducing an additional dimension of knowledge beyond data representations \cite{pop00001, pop00002, pop00004, pop00007, pop00008, pop00010, pop00011, pop00012}. In RE, each new model version is designed not only to learn from raw input data but also from the internal representations and estimated outputs generated by earlier iterations of the same architecture. This establishes a recurrent, reflective learning loop. Formally, RE model is expressed as in \autoref{eq:re_model}, where \( x \) is the raw input data, \( \phi(x) \) is the feature representation extracted by a deep neural architecture, and \( \tilde{y}_i \) is the estimated target produced by a previous model iteration \( f_{i-1} \). The function \( \rho[\phi(x), \tilde{y}_i] \) combines and transforms the internal feature maps and output predictions into a form that the current model \( f_i \) can use, resulting in a refined prediction \( y_i \).

\begin{equation}
y_i = f_i\left(x, \rho\left[\phi(x), \tilde{y}_i\right]\right)
\label{eq:re_model}
\end{equation}

We refer to the internal components \( \phi(x) \) and \( \tilde{y}_i \) as IMTs, standing for Input, Mappings, and estimated Targets. These elements capture both the internal state of the network and its prior output behavior. The role of the function \( \rho \) is critical in this context. Given the potentially high dimensionality of \( \phi(x) \), which may consist of deep feature maps with thousands of activation channels, \( \rho \) serves as a processing mechanism to reduce, project, or filter this information into a more compact and meaningful representation. This can be achieved through dimensionality reduction techniques, pooling strategies, learned projections, or attention-based mechanisms that focus on the most informative aspects of the combined signal. As shown in \autoref{eq:re_model}, this formulation enables recursive refinement by conditioning each model \( f_i \) on both the input and prior predictions. By integrating IMTs into the learning process, RE introduces a novel self-referential feedback loop. Each model becomes aware of the internal representations and outputs of its predecessor, allowing it to refine its understanding of the data and its own learning dynamics. This approach enables models to learn not only from data but also from the behavior of the learning process itself, introducing an additional, non-data-centric source of knowledge. RE thus provides a foundation for more adaptive, self-improving, and generalizable intelligence systems, pushing beyond the static, feedforward paradigm of traditional DL.

To monitor the quality and effectiveness of learning across RE iterations, we propose the use of the Area Under the Loss Curve (AULC) as a behavioral metric. The AULC provides a cumulative measure of how quickly and smoothly a model converges during training. In the context of RE, where models learn from both data and the traces of previous models, a lower and smoother AULC across iterations is indicative of more efficient learning. A decreasing AULC implies that the model is converging faster or with fewer fluctuations, suggesting that the knowledge transferred through IMTs is beneficial. This metric serves as a practical and interpretable tool to assess how well RE leverages prior learning behavior to enhance future learning performance, beyond what raw accuracy or loss at convergence alone can capture. Formally, we define AULC for a given model iteration \( i \) as in \autoref{eq:aulc}, where \( \mathcal{L}_i(t) \) is the loss of model \( f_i \) at training step \( t \), and \( T \) is the maximum number of training steps. This integral measures the accumulated loss over the course of training and reflects both the rate and stability of convergence. A lower AULC value indicates that the model is learning more efficiently, with faster convergence and fewer fluctuations in the loss. When used within the RE framework, a consistent decrease in AULC across model iterations provides evidence that each successive model is benefiting from the behavioral insights of its predecessor, as quantified by \autoref{eq:aulc}.

\begin{equation}
\mathrm{AULC}_i = \frac{1}{T} \int_{0}^{T} \mathcal{L}_i(t) \, dt
\label{eq:aulc}
\end{equation}

Algorithm \ref{alg:basic_re} provides a simplified pseudo-code representation of the core structure of the basic RE algorithm.

\begin{algorithm}
\caption{Recurrent Expansion Framework}
\label{alg:basic_re}
\begin{algorithmic}[1]
\Require Dataset $\mathcal{D} = \{(x, y)\}$, number of iterations $N$, model architecture $f$, feature extractor $\phi$, aggregator function $\rho$
\Ensure Final trained model $f_N$
\State Train base model $f_0$ on raw data $\mathcal{D}$
\For{$i = 1$ to $N$}
    \State Initialize recurrent-expanded dataset $\mathcal{D}_{\text{RE}} \gets \emptyset$
    \ForAll{$(x, y) \in \mathcal{D}$}
        \State $\phi_{\text{prev}} \gets \phi(x)$ from $f_{i-1}$
        \State $\tilde{y}_{i-1} \gets f_{i-1}(x)$
        \State $\text{IMT} \gets \rho(\phi_{\text{prev}}, \tilde{y}_{i-1})$
        \State $x_{\text{RE}} \gets (x, \text{IMT})$
        \State Append $(x_{\text{RE}}, y)$ to $\mathcal{D}_{\text{RE}}$
    \EndFor
    \State Train model $f_i$ on $\mathcal{D}_{\text{RE}}$
\EndFor
\State \Return $f_N$
\end{algorithmic}
\end{algorithm}

While RE introduces a promising mechanism for leveraging model behavior over time, it also introduces a unique vulnerability. Since each model in RE sequence builds on the internal representations and predictions of its predecessor, a poorly trained or misaligned model can inject inaccurate or distorted information into subsequent iterations. This phenomenon, which we refer to as a \textit{representation glitch} \cite{pop00006}, can significantly disrupt the learning process. In such cases, rather than observing a progressively smoother and decreasing AULC, the sequence may experience spikes or divergence, signaling that the transferred knowledge is degrading rather than enhancing performance. Addressing this challenge requires mechanisms that can detect or mitigate the propagation of flawed representations. One promising approach, inspired by ensemble and diversification strategies in machine learning, is the introduction of \textit{Multiverse Representations}, where multiple parallel models are trained during the initial stage instead of relying on a single prior. These models offer alternative behavioral trajectories and allow the system to either select or combine the most informative and stable sources of IMTs. In the following section, we explore MVRE and its variants as robust solutions to the representation glitch problem.

\section{Multiverse Recurrent Expansion}
\label{sec:mvre}

To address the potential instability in RE caused by dependence on a single previous model's behavior, we propose an extension termed MVRE \cite{pop00003, pop00005,pop00006, pop00009}. Rather than relying on a single predecessor to provide representational and behavioral feedback, MVRE maintains a set of parallel models from the previous iteration, each trained independently on the same data but initialized or regularized differently. These models form what we call the \textit{multiverse} that is a collection of diverse but structurally identical models that offer multiple perspectives on the learning process. By drawing from this multiverse, the current model can form a richer and more robust understanding of both the data and its past predictive landscape. Formally, in MVRE, the prediction at RE round \( i \) is given by \autoref{eq:mvre}.
\begin{equation}
y_i = f_i\left(x, \rho\left(\left\{\phi_j(x), \tilde{y}_j\right\}_{j=1}^{k}\right)\right)
\label{eq:mvre}
\end{equation}

Here, \( x \) is the input data, \( \phi_j(x) \) is the deep representation produced by the \( j \)-th model in the multiverse from RE round \( i-1 \), and \( \tilde{y}_j \) is its corresponding estimated target. The function \( \rho \) processes and combines this set of representations and predictions, indexed over \( j = 1, \ldots, k \), into a unified signal passed to the current model \( f_i \), which then produces the output \( y_i \). As in the single-model case, the function \( \rho \) plays a key role in reducing the complexity of the combined IMTs and highlighting salient information.

MVRE therefore generalizes RE paradigm by introducing diversity in the feedback mechanism. Instead of treating the previous model as a single authority, MVRE aggregates behavioral signals from a set of models trained in parallel, each potentially capturing different aspects of the learning space. This redundancy offers a safeguard against representational glitches by allowing the current model to smooth over outlier behaviors and emphasize consistent patterns across the multiverse. Furthermore, the diversity embedded in the multiverse introduces a form of ensemble-like robustness into the RE cycle, while still preserving the architecture-level recursion that characterizes RE. By iteratively applying this multiverse-informed learning across RE rounds, MVRE enables a more stable and adaptive learning process. It allows models not only to learn from data but also to build on an evolving ensemble of past representations and behaviors, ultimately advancing toward a form of introspective and collective learning.

In MVRE framework, the set of representations and predictions \(\{\phi_j(x), \tilde{y}_j\}_{j=1}^{k}\) is not passed directly to the next model. Instead, it is routed through a secondary deep network, designed specifically to interpret, compress, and evaluate the multiverse feedback. This auxiliary network acts as a connective module, learning how to transform the multiverse information into a form suitable for the current model iteration \( f_i \). Importantly, this network also plays a critical role in assessing the quality of each source via behavioral indicators such as AULC. By monitoring AULC of each prior model, the system can dynamically weigh, filter, or prioritize certain representations during aggregation. This allows MVRE not only to integrate multiple behavioral signals but to adaptively learn which historical trajectories were more reliable, thereby preventing the propagation of suboptimal learning patterns. As a result, the recurrent loop in MVRE becomes both behavior-aware and selectively introspective, using an additional level of abstraction to refine how past knowledge informs future learning.

The pseudocode in Algorithm~\ref{alg:mvre} outlines the core structure of MVRE algorithm.

\begin{algorithm}[ht]
\caption{Multiverse Recurrent Expansion}
\label{alg:mvre}
\begin{algorithmic}[1]
\Require Dataset $\mathcal{D}$, number of rounds $N$, multiverse size $k$
\Ensure Final model $f_N$

\State Initialize base models $\{f^0_j\}_{j=1}^k$ independently
\For{$j = 1$ to $k$}
    \State Train $f^0_j$ on $\mathcal{D}$
    \State Store representations $\phi^0_j(x)$ and predictions $\tilde{y}^0_j$
\EndFor

\For{$i = 1$ to $N$}
    \State Build multiverse set $\mathcal{M}^{i-1} = \{(\phi^{i-1}_j(x), \tilde{y}^{i-1}_j, \mathrm{AULC}^{i-1}_j)\}_{j=1}^k$
    \State Aggregate IMTs via auxiliary module: $\rho_i = \text{Aggregate}(\mathcal{M}^{i-1})$
    \State Train current model $f_i$ using inputs $(x, \rho_i)$ and labels from $\mathcal{D}$
    \State Compute predictions $y_i = f_i(x, \rho_i)$

    \For{$j = 1$ to $k$}
        \State Initialize $f^i_j$ with random seed or regularization
        \State Train $f^i_j$ on $\mathcal{D}$
        \State Store $\phi^i_j(x)$, $\tilde{y}^i_j = f^i_j(x)$, and $\mathrm{AULC}^i_j$
    \EndFor
\EndFor

\State \Return $f_N$
\end{algorithmic}
\end{algorithm}

\section{Heterogeneous Multiverse Recurrent Expansion}
\label{sec:hmvre}

Despite its enhanced robustness, MVRE as described thus far still assumes a homogeneous pool of models, that is, all multiverse models share the same architecture. This constraint, while simplifying integration, limits the diversity of representational features and behavioral dynamics available to the RE process. Moreover, if the parallel models in the multiverse are poorly selected or collectively biased, the aggregated information may still propagate suboptimal behavior into future iterations. However, the representation glitch is still mitigated to some extent, as the presence of multiple models ensures that outliers are less likely to dominate. 

To further strengthen this robustness and increase representational diversity, we introduce an extension: HMVRE) \cite{pop00006}. In HMVRE, the multiverse pool is composed of structurally different model types. For instance, CNNs, recurrent models such as LSTMs, GNNs, and transformer-based encoders. Each of these architectures processes the input \( x \) in its unique inductive bias and abstraction mechanism, providing complementary perspectives on the same data.

Formally, HMVRE can be expressed as \autoref{eq:hmvre}.

\begin{equation}
y_i = f_i\left(x, \rho\left(\left\{\phi_j^{(a_j)}(x), \tilde{y}_j\right\}_{j=1}^{k}\right)\right)
\label{eq:hmvre}
\end{equation}

where \( \phi_j^{(a_j)}(x) \) denotes the feature representation of input \( x \) produced by the \( j \)-th model of architecture type \( a_j \), and \( \tilde{y}_j \) is its corresponding predicted output. The function \( \rho \) again serves to process and integrate these heterogeneous representations and behavioral traces into a unified, learnable signal for the current model \( f_i \). By leveraging architectural diversity, HMVRE enables richer multiview representation learning and enhances the system’s resilience to representational collapse or uniform error propagation. This extension moves RE framework closer to the goal of generalized and introspective learning by capturing structural, sequential, and relational aspects of the data through complementary inductive pathways.

The pseudocode in Algorithm \ref{alg:hmvre} outlines the structure of HMVRE algorithm, which integrates multiple model architectures to enhance representational diversity and robustness.

\begin{algorithm}[ht]
\caption{Heterogeneous Multiverse Recurrent Expansion}
\label{alg:hmvre}
\begin{algorithmic}[1]
\Require Dataset $\mathcal{D}$, number of rounds $N$, heterogeneous model set $\{a_j\}_{j=1}^k$
\Ensure Final model $f_N$

\For{$j = 1$ to $k$}
    \State Initialize model $f^0_j$ of type $a_j$
    \State Train $f^0_j$ on $\mathcal{D}$
    \State Store $\phi^{0,(a_j)}_j(x)$ and $\tilde{y}^0_j$
\EndFor

\For{$i = 1$ to $N$}
    \State Form heterogeneous multiverse set:
    \[
    \mathcal{M}^{i-1} = \{(\phi^{i-1,(a_j)}_j(x), \tilde{y}^{i-1}_j)\}_{j=1}^k
    \]
    \State Compute aggregated signal: $\rho_i = \text{Aggregate}(\mathcal{M}^{i-1})$
    \State Train current model $f_i$ on $(x, \rho_i)$
    \State Predict $y_i = f_i(x, \rho_i)$

    \For{$j = 1$ to $k$}
        \State Initialize new $f^i_j$ of architecture $a_j$
        \State Train $f^i_j$ on $\mathcal{D}$
        \State Store $\phi^{i,(a_j)}_j(x)$ and $\tilde{y}^i_j$
    \EndFor
\EndFor

\State \Return $f_N$
\end{algorithmic}
\end{algorithm}

\section{Scalable Heterogeneous Multiverse Recurrent Expansion}
\label{sec:sc_hmvre}

To further generalize and adapt the RE paradigm for practical deployment at scale, we propose Sc-HMVRE. This formulation extends HMVRE by incorporating models of varying scales, from lightweight, low-latency architectures to large-scale, high-capacity deep models, into a unified multiverse framework. By integrating models across a spectrum of computational and representational capacities, Sc-HMVRE supports both efficiency and expressiveness, making it suitable for diverse environments ranging from edge devices to cloud infrastructure. In addition to scale diversity, Sc-HMVRE introduces a selective mechanism that chooses which subset of the heterogeneous multiverse to engage at each RE iteration. This selection can be based on multiple criteria such as computational budget, prior AULC scores, domain-specific relevance, or learned importance weights. This allows the system to balance exploration of multiple learning pathways with exploitation of reliable prior models.

Formally, the Sc-HMVRE model at RE round \( i \) is expressed as in \autoref{eq:sc_hmvre}.

\begin{equation}
y_i = f_i\left(x, \rho\left(\left\{ \phi_j^{(a_j, s_j)}(x), \tilde{y}_j \right\}_{j \in \mathcal{S}_i} \right)\right)
\label{eq:sc_hmvre}
\end{equation}

Here, \( \phi_j^{(a_j, s_j)}(x) \) represents the feature mapping from the \( j \)-th model of architecture type \( a_j \) and scale \( s_j \) (e.g., small, medium, large). The index set \( \mathcal{S}_i \subseteq \{1, \ldots, k\} \) denotes the selected subset of multiverse models used at iteration \( i \), chosen based on selection criteria described above. As before, \( \tilde{y}_j \) is the corresponding target estimate, and \( \rho \) fuses the behavioral traces and representations into an integrated signal for learning.

Sc-HMVRE provides a flexible and adaptive framework for real-world learning systems. It enables the combination of small-scale models, which offer speed and adaptability, with large-scale models, which capture complex structures and patterns. Moreover, the selective mechanism introduces dynamic control over computational load and learning diversity. This results in a scalable RE strategy capable of continuous learning in resource-aware, multi-environment, and behavior-sensitive settings.

Algorithm~\ref{alg:sc_hmvre} presents Sc-HMVRE, which extends HMVRE with scale diversity and a dynamic selection mechanism for improved efficiency and adaptability.

\begin{algorithm}[ht]
\caption{Scalable Selective Heterogeneous Multiverse Recurrent Expansion}
\label{alg:sc_hmvre}
\begin{algorithmic}[1]
\Require Dataset $\mathcal{D}$, number of rounds $N$, heterogeneous-scale model set $\{(a_j, s_j)\}_{j=1}^k$
\Ensure Final model $f_N$

\For{$j = 1$ to $k$}
    \State Initialize model $f^0_j$ with architecture $a_j$ and scale $s_j$
    \State Train $f^0_j$ on $\mathcal{D}$
    \State Store $\phi^{0,(a_j, s_j)}_j(x)$, $\tilde{y}^0_j$, and AULC$_j^0$
\EndFor

\For{$i = 1$ to $N$}
    \State Select subset $\mathcal{S}_i \subseteq \{1, \ldots, k\}$ based on AULC, budget, or relevance
    \State Build selected multiverse set:
    \[
    \mathcal{M}_i = \{(\phi^{i-1,(a_j, s_j)}_j(x), \tilde{y}^{i-1}_j)\}_{j \in \mathcal{S}_i}
    \]
    \State Aggregate: $\rho_i = \text{SelectiveAggregate}(\mathcal{M}_i)$
    \State Train model $f_i$ on $(x, \rho_i)$
    \State Predict $y_i = f_i(x, \rho_i)$

    \For{$j = 1$ to $k$}
        \State Initialize $f^i_j$ with $(a_j, s_j)$
        \State Train $f^i_j$ on $\mathcal{D}$
        \State Store $\phi^{i,(a_j, s_j)}_j(x)$, $\tilde{y}^i_j$, and AULC$_j^i$
    \EndFor
\EndFor

\State \Return $f_N$
\end{algorithmic}
\end{algorithm}

\section{Illustrative Example: Sinusoidal Regression via Recurrent Expansion}
\label{sec:example}

To demonstrate the behavior and performance of the proposed RE framework, we present a synthetic regression task based on a sinusoidal target function. The input \( x \in \mathbb{R}^{n \times 1} \) consists of 100 uniformly spaced scalar values, and the output \( y \in \mathbb{R}^{n \times 1} \) is generated according to the noisy sine model described in Equation~\eqref{eq:sinusoidal_model}. This setup provides a clear, nonlinear regression task suitable for analyzing the effect of RE on representational learning.

\begin{equation}
    y = \sin(2\pi x) + \epsilon, \quad \epsilon \sim \mathcal{N}(0, \sigma^2)
    \label{eq:sinusoidal_model}
\end{equation}

\begin{figure}[ht]
    \centering
    \includegraphics[width=0.6\linewidth]{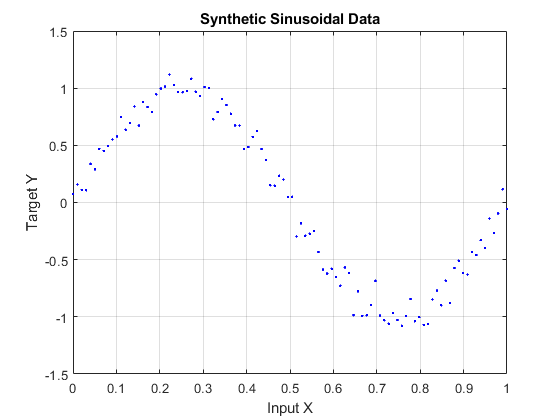}
    \caption{Synthetic sinusoidal dataset used for evaluating the Recurrent Expansion framework.}
    \label{fig:dataset_example}
\end{figure}

The characteristics of the generated sinusoidal dataset used in this study are visualized in Figure~\ref{fig:dataset_example}.

The goal is to learn the underlying mapping using Multilayer Perceptron (MLP)-based RE model with Principal Component Analysis (PCA)-compressed internal representations. The retained variance threshold for PCA was set to 20\%, and RE process was executed for 100 iterations.

\begin{figure}[ht]
    \centering
    \includegraphics[width=0.6\linewidth]{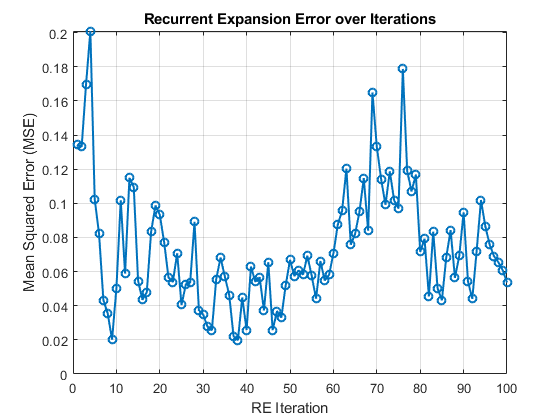}
    \caption{Mean Squared Error (MSE) across RE iterations for the sinusoidal regression task.}
    \label{fig:re_error}
\end{figure}

As shown in Figure~\ref{fig:re_error}, RE framework demonstrates the ability to progressively refine the model through self-reflective learning. Despite some fluctuations in the early rounds, the error notably drops from approximately 0.20 at iteration 4 to a minimum of 0.01 at iteration 38. This confirms the model's ability to iteratively improve its internal representation and prediction accuracy.

Interestingly, we observe a phenomenon we refer to as \textit{representation glitch}, where the error begins to rise again after the initial convergence. This increase may indicate that the accumulated representation or prediction information starts to interfere destructively with learning, suggesting a need for the aformentioned MV paradigms for further improvements or early stopping in practical implementations. The MATLAB implementation used for this example is provided in the Code Availability Section.

\section{Future Directions}
\label{sec:related}

The RE framework and its multiverse extensions open up a new paradigm in DL (i.e., one that integrates not only data-driven learning but also learning from the behavior and evolution of models themselves). Moving forward, this approach holds the potential to reshape how DL systems are trained, evaluated, and deployed. First, future work can explore more advanced mechanisms for integrating model behavior, beyond simple representations and predictions. For example, learning dynamics such as gradient flows, activation patterns, or attention distributions may serve as rich behavioral descriptors. By incorporating these signals into RE cycle, models could learn not only what they predict, but how they reach those predictions.

Second, the introduction of heterogeneous and scalable multiverse models in HMVRE and Sc-HMVRE invites investigation into curriculum-based and architecture-aware selection strategies. Dynamic scheduling algorithms could prioritize which models are most informative at different stages of learning, based on task complexity, uncertainty, or resource availability. This selective introspection could lead to more efficient and adaptive lifelong learning systems. Third, the auxiliary networks used for aggregation and AULC evaluation could themselves become trainable meta-models, enabling end-to-end optimization of RE process. By learning how to interpret and compress behavioral traces, these modules may unlock new forms of meta-learning and self-diagnosis, driving improvements not only in accuracy but in learning efficiency, robustness, and generalization.

Finally, RE opens the door to a deeper form of self-aware artificial intelligence, where models build a longitudinal memory of their own experiences and performance histories. This could pave the way for new learning systems that reason not only over static data but over the temporal structure of their own evolution. As RE frameworks mature, they may lead to transformative shifts in how we conceptualize and design intelligent systems, moving from static, task-specific learners to dynamic, reflective, and behavior-sensitive agents capable of continual adaptation and knowledge accumulation. In this way, RE is not just a method, in fact, it is a shift in how learning itself is modeled, evaluated, and extended.

\section{Conclusion}
\label{sec:conclusion}

This work introduced the concept of RE as a novel learning paradigm that shifts the focus of DL beyond traditional data-driven training. By learning not only from data representations but also from model behaviors and performance histories, RE offers a new axis of knowledge previously overlooked in DL research. Through the extensions MVRE, HMVRE, and Sc-HMVRE, we demonstrated how multiverse structures, architectural heterogeneity, and scalable selection mechanisms can be integrated into a unified framework for continual behavioral learning. These approaches enrich the representational landscape by capturing diverse, behavior-aware trajectories through deep networks, allowing models to iteratively refine their understanding of both data and learning dynamics. The inclusion of auxiliary networks and meta-behavioral evaluations such as AULC provides additional insights into the quality of learning, offering new tools for diagnostics, optimization, and self-improvement. RE marks a step toward self-reflective DL systems that do not merely learn to fit data but evolve through experience and introspection. As DL continues to scale in complexity and deployment, such mechanisms for behavioral awareness and multiverse reasoning may become essential. RE, therefore, represents not just a methodological extension, but a foundational direction toward a more general, adaptive, and cognitively inspired future of machine learning.

\section*{Declarations}

\paragraph*{Funding:} This research received no external funding.

\paragraph*{Conflict of Interest:} The author declares no conflict of interest.

\paragraph*{Ethical Approval:} Not applicable.

\paragraph*{Consent to Participate:} Not applicable.

\paragraph*{Consent for Publication:} Not applicable.

\paragraph*{Availability of Data and Materials:} 
The synthetic dataset used in this study is available at: \href{https://doi.org/10.5281/zenodo.15808571}{https://doi.org/10.5281/zenodo.15808571}.

\paragraph*{Code Availability}
\label{par:code_availability}
The code used for the experiments and figures in this work is available at: \href{https://doi.org/10.5281/zenodo.15808571}{https://doi.org/10.5281/zenodo.15808571}.

\paragraph{Author Contributions:} The author is solely responsible for all aspects of the manuscript.

\bibliographystyle{unsrt}  
\bibliography{references}  
\end{document}